\documentclass[11pt,a4paper]{article}
\usepackage{times}
\usepackage{latexsym}
\usepackage{amsmath}
\usepackage{amsfonts}
\usepackage{graphicx}
\usepackage{multirow}
\usepackage{color}
\usepackage{subfigure}
\usepackage{microtype}

\title{Downstream Model Design of Pre-trained Language Model for Relation Extraction Task}

\author{Cheng Li, Ye Tian \\
 AI Application Research Center, 
 Huawei Technologies,
  Shenzhen, China \\
  \texttt{licheng81@huawei.com} }

\date{}

\begin{document}
\maketitle
\begin{abstract}
Supervised relation extraction methods based on deep neural network play an important role in the recent information extraction field. However, at present, their performance still fails to reach a good level due to the existence of complicated relations. On the other hand, recently proposed pre-trained language models (PLMs) have achieved great success in multiple tasks of natural language processing through fine-tuning when combined with the model of downstream tasks. However, original standard tasks of PLM do not include the relation extraction task yet. We believe that PLMs can also be used to solve the relation extraction problem, but it is necessary to establish a specially designed downstream task model or even loss function for dealing with complicated relations. In this paper, a new network architecture with a special loss function is designed to serve as a downstream model of PLMs for supervised relation extraction. Experiments have shown that our method significantly exceeded the current optimal baseline models across multiple public datasets of relation extraction.
\end{abstract}

\section{Introduction}
As a subtask of information extraction, the predefined relation extraction task plays an important role in building structured knowledge. In recent years, with the rapid development of deep learning, many relation extraction methods based on deep neural network structure have been proposed. At present, these popular neural network methods~\cite{liu2013convolution,zeng2014relation,nguyen2015relation,lin2016neural,jiang2016relation,han2018hierarchical,fu2019graphrel,soares2019matching} can be mainly summarized by the following steps:

1. Obtain embeddings of the target text from an encoder. The embeddings are usually dense vectors in a low dimensional space~\cite{mikolov2013distributed}. For example, Glove vector~\cite{pennington2014glove}, or vectors from a pre-trained language model like BERT~\cite{devlin2018bert}.

2. Process these embeddings and integrate their information according to a certain network structure, such as CNN~\cite{nguyen2015relation}, Attention Mechanism~\cite{lin2016neural,han2018hierarchical} and GCN~\cite{fu2019graphrel} and so on, to obtain the representation of the target relation.

3. Perform training on labeled dataset based on a certain classifier using the encoded information as input, such as the Softmax classifier.

However, the current methods do not perform very well on public datasets given the existence of complicated relations, such as long-distance relation, single sentence with multiple relations, and overlapped relations on entity-pairs.

Recently, the emergence of pre-trained language models has provided new ideas for solving relation extraction problems. Pre-trained language models~\cite{devlin2018bert,radford2019language,yang2019xlnet,raffel2019exploring} are a kind of super-large-scale neural network model based on the deep Transformer~\cite{vaswani2017attention} structure. Their initial parameters are learned through super-large self-supervised training, and then combined with multiple downstream models to fix special tasks by fine-tuning. Experiments show that the downstream tasks' performances of this kind of models are usually far superior to those of conventional models~\cite{liu2019roberta}. Unfortunately, the relation extraction task is not in the original downstream tasks list directly supported by the pre-trained language model.

Our current work attempts to leverage the power of the PLMs to establish an effective downstream model that is competent for the relation extraction tasks. In particular, we implement \textbf{three important improvements} in the main steps, as described above:

1. Use a pre-trained language model (this article uses BERT~\cite{devlin2018bert}) instead of the traditional encoder, and obtain \textbf{a variety of token embeddings} from the model. We extract the embeddings from two different layers to represent the head and tail entities separately, because it may help to learn reversible relations. On the other hand, we also add the context information into the entity to deal with long-distance relations.

2. After then, we calculate a \textbf{parameterized asymmetric kernel inner product matrix} between all the head and tail embeddings of each token in a sequence. Since kernels are different between each relation, we believe such a product is helpful for distinguishing multiple relations between same entity pairs. Thus the matrix can be treated as the tendency score to indicate where a certain relation exists.

3. Use \textbf{the Sigmoid classifier instead of the Softmax classifier}, and use the average probability of token pairs corresponding to each entity pair as the final probability that the entity pair has a certain relation. Thus for each type of relation, and each entity pair in the input text, we can independently calculate the probability of whether a relation exists. This mechanism allows the model to predict multiple relations even overlapped on same entities.

We also notice that it is very easy to integrate Named Entity Recognition (NER) task into our model to deal with joint extraction task. For instance, adding Bi-LSTM and CRF~\cite{huang2015bidirectional} after the BERT encoding layer makes it easy to build an efficient NER task processor. By simply adding the NER loss on original loss, we can build a joint extraction model to avoid the pipeline errors. However, in this paper we will focus on relation extraction and will not pay attention to this method.

Our experiments mainly verify two conclusions.

1. Pre-trained language model can perform well in relation extraction task after our re-designing. We evaluate the method on $3$ public datasets: SemEval 2010 Task 8~\cite{hendrickx2009semeval}, NYT~\cite{riedel2010modeling}, WebNLG~\cite{gardent2017creating}. The experimental results show that our model achieves a new state-of-the-art.

2. A test for our performance in different overlapped relation situations and multiple relation situations shows that our model is robust when faced with complex relations.

\section{Related Work}
In recent years, numerous models have been proposed to process the supervised relation extraction task with deep neural networks. The general paradigm is to create a deep neural network model by learning from texts labeled with entities information and the relation-type-label between them. The model then can predict the possible relation-type-label for the specified entities in a new text. We introduce three kinds of common neural network structures here.

\subsection{CNN-based methods} CNN based methods apply Convolutional Neural Network~\cite{lecun1998gradient} to capture text information and reconstruct embeddings~\cite{liu2013convolution}. Besides simple CNN, a series of improved models have been proposed~\cite{liu2013convolution,zeng2014relation,nguyen2015relation,lin2016neural,jiang2016relation}. However, the CNN approach limits the model's ability to handle remote relations. The information that is fused through the CNN network is often local, so it is difficult to deal with distant relations. Therefore, these methods are currently limited to a large extent and are not able to achieve a good level of application. Since more efficient methods are proposed recently, we will not compare our approach with this early work in this article.

\subsection{GNN-based methods} GNN based methods use Graph Neural Network~\cite{scarselli2008graph}, mainly Graph Convolutional Network~\cite{kipf2016semi} to deal with entities' relations. GNN is a neural network that can capture the topological characteristics of the graph type data, so it is necessary to define a prior graph structure. The GNN-based relation extraction methods~\cite{zhang2018graph,guo2019attention,fu2019graphrel} usually use the text dependency tree as an input prior graph structure, thereby obtaining richer information representation than CNN. However, this kind of methods relies on the dependency parser thus pipeline errors exist. In addition, grammar-level dependency tree is still a shallow representation which fails to efficiently express relations between words. 

\subsection{Methods with Pre-trained Language Model}
Some recent approaches consider relation extraction as a downstream task for PLMs. These methods~\cite{zhang2019ernie,soares2019matching,joshi2019spanbert,wei2019novel} have made some success, but we believe that they have not yet fully utilized the language model. The main reason is the lack of a valid representation of the relation - those methods tend to express the relation as a one-dimensional vector. We believe that since the relation is determined by the vectors' correlations of head and tail entities, it should naturally be represented as a matrix rather than one-dimensional vector. In this way, more information like the order of the entities and their positions in the text will be used while predicting their relations.

\begin{figure*}[ht]

\centering
\includegraphics[scale=0.22]{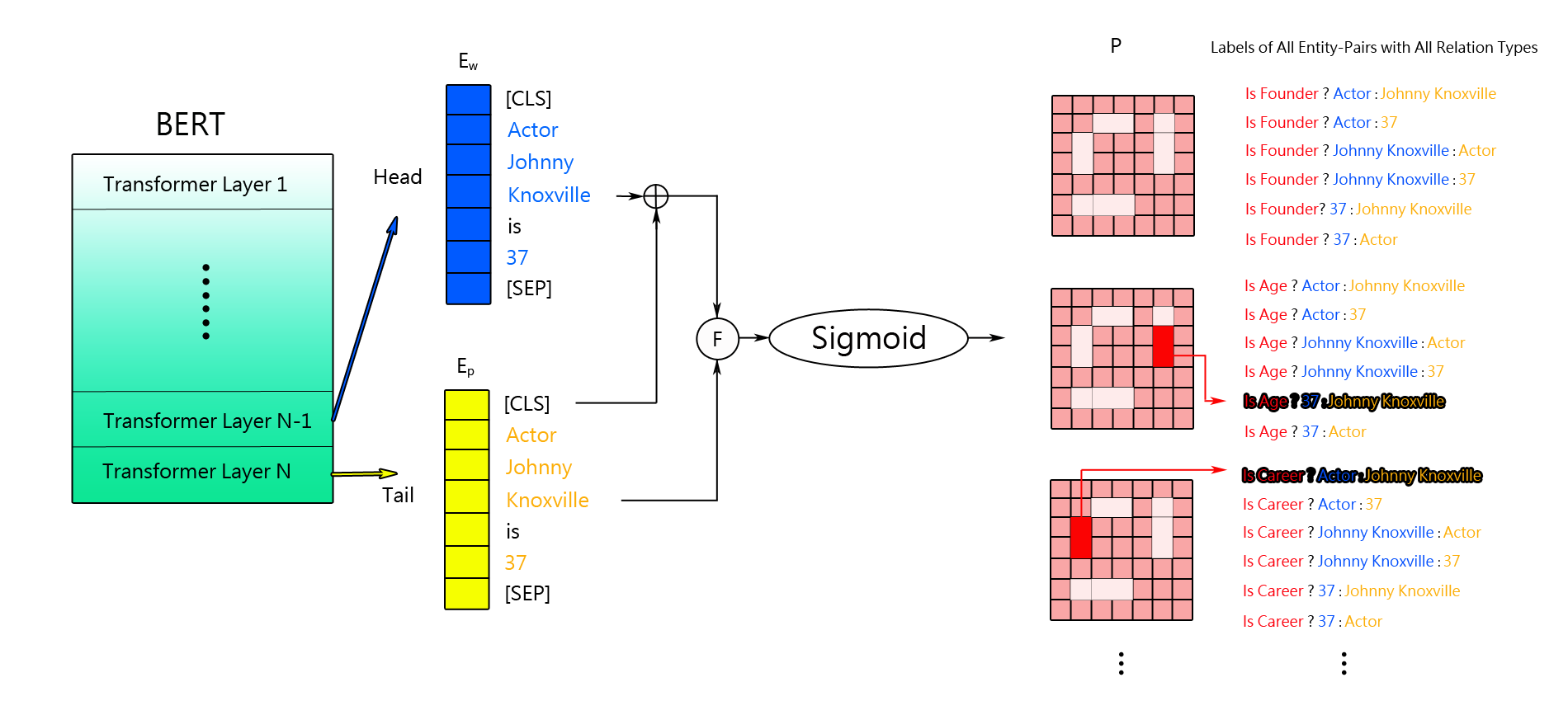}
\caption{Our method's network architecture. $F$ is an asymmetric kernel inner product function of embeddings at each position. Thus we get a product matrix of $l\times l$ ($l=7$ in this figure) for each relation type. We further use Sigmoid activation function to scale each element and get probability matrix $P$. For each relation type, the average value of the elements that correspond to the entity-pair can be treated as the final predicted score for the possible triplet $\left \langle relation ? head : tail\right \rangle$.}
\label{netfig}
\end{figure*}

\section{Methodology}
\label{method}

We introduce the method from two aspects: network structure and loss function.
Figure~\ref{netfig} shows the overall architecture of this method.

From the perspective of the network structure, our model has two major parts. The first part is an encoder that utilizes pre-trained language model like BERT. We obtain three embeddings for a given input text from the BERT model: the embedding $\boldsymbol{E}_w$  for each token in the text , the embeddings $\boldsymbol{E}_p$ obtained by passing  $\boldsymbol{E}_w$ through a self-attention Transformer, and the embedding $\boldsymbol{E}_a$ of the entire text (that is, the CLS embedding provided by BERT). The second part is the relation computing layer. In this layer, we assume that the $\boldsymbol{E}_p$  represents the tail entity encoded with some available predicate information, while $\boldsymbol{E}_a$ combined with $ \boldsymbol{E}_w$  represents the head entity with context information. By performing a correlation calculation $F$ on those embeddings, the tendency scores matrix $\boldsymbol{S}_i$ of relation $i$ in all entity pairs can be obtained.

From the perspective of loss function, we first use the Sigmoid activation function to compute probabilites $\boldsymbol{P}_i$ of relation $i$ by using $\boldsymbol{S}_i$. We use locations of entities to construct a mask matrix $\boldsymbol{M}$ and use it to preserve information in $\boldsymbol{P}_i$ which represents existing entity-pairs in a sentence (See details in Section~\ref{loss} and Figure~\ref{netfig}). Based on the labels indicating whether each entity-pair is an instance of the $i$-th relation type or not, we use the average values in each area of $\boldsymbol{P}_i$ to compute a Binary Cross Entropy (BCE) loss of this specific relation. Eventually, the final loss sums all values from all relations. This formulation allows the model to predict multiple relations in a single sentence or even multiple relations for a single entity-pair.
Details and formulas are described in subsections below.

\subsection{Encoder}
The current pre-training language models are basically based on the Transformer structure~\cite{vaswani2017attention} . They have a common feature: each layer of the Transformer can output a hidden vector corresponding to the input text $T$. $T$ is an array of tokens with length $l$. Taking BERT as an example, some simple analyses ~\cite{devlin2018bert} show that the hidden vector of each layer can be used as word embeddings of $T$, with modest difference in precisions. Generally speaking, the deepest hidden vector representation of the Transformer network tends to work best for a downstream fine-tuning task thanks to the information integration performed by a deeper network. However, here we select the penultimate layer output vector as the initial embedding $\boldsymbol{E}_w$ ($\boldsymbol{E}_w\in \mathbb{R}^{l\times h}$, where $h$ is the number of hidden dimensions), for the text representation with entity information.

To get $\boldsymbol{E}_p$, we use the last Transformer layer of BERT which is actually a multi-head self-attention with a fully connected FFN~\cite{vaswani2017attention} to deal with our initial embeddings:

\begin{equation}
    \boldsymbol{E}_{p}=Transformer(\boldsymbol{E}_w),
\end{equation}
where $\boldsymbol{E}_p$ is the last output vector of BERT, $\boldsymbol{E}_p \in \mathbb{R}^{l\times h}.$

Such an operation is applied so that the embedding of every token in $\boldsymbol{E}_p$ will, in addition to $\boldsymbol{E}_w$, fuse some information from tokens in other positions. In this way, Although dataset annotations usually do not carry explicit predicate information, the Transformer structure of the BERT model allows  $\boldsymbol{E}_p$ to selectively blend contextual information that is helpful for the final task. We expect that after well fine-tuning training, words with higher attention association scores correspond to a predicate of a certain relation to some extent. $\boldsymbol{E}_w$ and $\boldsymbol{E}_p$ can be respectively used as basic entity representations and entity representations that incorporate predicate information. In order to better capture the overall context information, the BERT's CLS embedding $\boldsymbol{E}_a$($\boldsymbol{E}_a\in \mathbb{R}^{h}$) is also added to each token's embedding to improve the basic entity representation:
\begin{equation}
    \boldsymbol{E}_{b}=\boldsymbol{E}_w+\boldsymbol{E}_a.
\end{equation}
Note $\boldsymbol{E}_a$ is actually broadcasting to all tokens.
\subsection{Relation Computing Layer}
We apply an asymmetric kernel inner product method to calculate the similarity between $\boldsymbol{E}_b$ and $\boldsymbol{E}_p$:
\begin{equation}
    \boldsymbol{S}_i=F_i(\boldsymbol{E}_b,\boldsymbol{E}_p),
\end{equation}
    where  
\begin{equation}F_i(\boldsymbol{X},\boldsymbol{Y})= \boldsymbol{X}\boldsymbol{W}_{hi}\cdot(\boldsymbol{Y}\boldsymbol{W}_{ti})^T.
\end{equation}
Here $\boldsymbol{S}_i\in \mathbb{R}^{l\times l};\boldsymbol{W}_{hi},\boldsymbol{W}_{ti}\in \mathbb{R}^{h\times h}$.

Actually, $\boldsymbol{W}_{hi}$ and $\boldsymbol{W}_{ti}$ are respectively the transformation matrices of head-entity and tail-entity embeddings in $i$-th relation. They are the parameters learned during the training process for each relation. 

If there are $N$ tokens in one input text, we find that $S_i$ is actually a square matrix with $N$ rows and columns. Thus it can be treated as unnormalized probability scores for $i$-th relation between all the tokens. That is to say, $\boldsymbol{S}_i\,_{mn}$, an element of position $(m,n)$, represents the existence possibility of $i$-th relation between tokens at these two locations. Finally we use Sigmoid functions to normalize $\boldsymbol{S}_i$ to range $(0,1)$:

\begin{equation}
    \boldsymbol{P}_{i}=\frac{1}{1+e^{-\boldsymbol{S}_i}}\,,
\end{equation}
where $\boldsymbol{P}_{i}$ is the normalized probability matrix of $i$-th relation. 

\subsection{Loss Calculation}
\label{loss}
A problem of $\boldsymbol{P}_i$ is that it describes relations between tokens, not entities. Therefore, we use entity-mask matrix to fix this problem. For each entity pair, the location information of the entities is known. Suppose that all entities from input text $T$ constitute a set of entity pairs in the form:$$\mathbb{S}=\{(x,y)\}.$$
Suppose $(B_x,E_x)$ is the beginning and end of the position index of an entity $x$ in the token array. Therefore, we construct a mask matrix $\boldsymbol{M}(\boldsymbol{M}\in\mathbb{R}^{l\times l})$ to satisfy
\begin{equation}
\begin{small}
     \forall (x,y) \in \mathbb{S}, \boldsymbol{M}_{mn}=\left\{
\begin{aligned}
&1, B_x\leq m\leq E_x \land B_y\leq n\leq E_y \\
&0, otherwise 
\end{aligned}
\right.
\end{small}
\end{equation}
where $m,n$ is the subscript of the matrix element. Similarly, we can construct a label matrix $\boldsymbol{Y_i}(\boldsymbol{Y_i}\in\mathbb{R}^{l\times l})$ for the $i$-th relation:
\begin{equation}
\begin{small}
     \forall(x,y)\in \mathbb{Y}_i, \boldsymbol{Y_i}_{mn}=\left\{
\begin{aligned}
&1, B_x\leq m\leq E_x \land B_y\leq n\leq E_y \\
&0, otherwise 
\end{aligned}
\right.
\end{small}
\end{equation}
where $\mathbb{Y}_i$ is the labeled $i$-th relation set of entity pairs from the input text $T$.
We use this mask matrix to reserve the predicted probabilities of every entity pair from $\boldsymbol{P}_i$, and then use the average Binary Cross Entropy to calculate the target loss $L_i$ of relation $i$:
\begin{equation}
    L_i=BCE_{avg}(\boldsymbol{P}_i*\boldsymbol{M},\boldsymbol{Y}_i)
\end{equation}
where $*$ is Hadamard product and
\begin{equation}
\begin{small}
\begin{aligned}
    &BCE_{avg}(\boldsymbol{X},\boldsymbol{Y})\\
    &=\frac{\sum\limits_{mn, \forall \boldsymbol{Y}_{mn}=1} log(\boldsymbol{X}_{mn})+\sum\limits_{mn, \forall \boldsymbol{Y}_{mn}=0} log(1-\boldsymbol{X}_{mn}) }{\sum\limits_{mn, \forall {\boldsymbol{M}_{mn}=1}}\boldsymbol{Y}_{mn}}
\end{aligned}
\end{small}
\end{equation}
Thus the final loss $L_r$ of relation predication is
\begin{equation}
\begin{aligned}
    L_r=\sum_{i}L_i
\end{aligned}
\end{equation}
where $i$ is the index of each relation. While predicting, we use the average value of elements in $\boldsymbol{P}_i$, whose location accords with a certain entity-pair $(x,y)$, as the probability of the possible triplet $\left \langle i?x:y \right \rangle $ consisting of $i$-th relation and entity-pair $(x,y)$.

\begin{table*}
\centering
\begin{tabular}{lccc}
\hline \textbf{Hyper-parameters} & \textbf{SemEval} & \textbf{NYT}&  \textbf{WebNLG}\\ \hline
Batch size   & 64 & 20 & 20 \\
Learning Rate & $3\times 10^{-5}$ & $5\times 10^{-5}$ &  $3\times 10^{-5}$\\
Maximum Training Epochs& 50 & 10 & 30 \\
Maximum Sequence Length & 512 & 100  & 512 \\
\hline
\end{tabular}
\caption{\label{para-tab} Hyper-parameters used for training on each dataset. }
\end{table*}
 \begin{table*}
\centering
\begin{tabular}{lccc}
\hline \textbf{Situations} & \textbf{SemEval} & \textbf{NYT}& \textbf{WebNLG}\\ \hline

Normal  & 10695 & 33566 ‬& 12391  \\
EPO  & 0 & 30775  & 121  \\
SEO & 0 & 13927 & 19059 \\
\hline
Single  & 10673 & 33287  & 12237  \\
Double  & 22 & 24174  & 9502  \\
Multiple & 0 & 12249  & 9772 \\
\hline
All & 10695 & 69710   & 31511 \\
\hline
\end{tabular}
\caption{\label{data-table} Statistics of different types of sample sentences (No repetition) in multiple datasets. Note a sample sentence may belong to both EPO and SEO. }
\end{table*}
\begin{table*}
\centering
\begin{tabular}{lcccc}
\hline \textbf{Methods} & \textbf{SemEval} & \textbf{NYT}& \textbf{WebNLG}\\ \hline
C-AGGCN~\cite{guo2019attention}   & 85.7 & -- & -- \\
GraphRel2p~\cite{fu2019graphrel} & -- & 61.9 & 42.9 \\
\hline
${\rm {BERT_{EM}}}$-MTB ~\cite{soares2019matching}& 89.5 & -- & --\\
HBT~\cite{wei2019novel} & -- & 87.5& 88.8 \\
\hline
ours & \textbf{91.0} & \textbf{89.8}  & \textbf{96.3}  \\
\hline
\end{tabular}
\caption{\label{preformance-table} Micro-F1 scores of our method tested on multiple datasets, compared with other four baseline methods. The first two methods are GNN-based while the last two are PLM-based. Their performances come from their original papers, as quoted above. }
\end{table*}
\begin{table*}
\centering
\begin{tabular}{lccccccccc}
\hline
\multirow{2}*{\textbf{Situations}}& \multicolumn{3}{c}{\underline{\textbf{SemEval}}}  & \multicolumn{3}{c}{\underline{\textbf{NYT}}}& \multicolumn{3}{c}{\underline{\textbf{WebNLG}}}\\
~& P& R &F1& P& R &F1& P& R &F1\\

\hline

Normal & 94.2 & 88.0 &91.0                 &95.1 & 94.0 & 94.5                         & 96.2 &92.9  & 94.5\\
EPO  & - & - &-               & 96.3& 73.2 & 83.2                        &100.0  & 90.3& 94.9 \\
SEO& - & - &-              & 92.2 & 78.9   & 85.0                          & 97.2 &96.3  & 96.8  \\
\hline
Single &94.2 &88.0  &  91.0                & 95.1& 94.0 & 94.6                         & 96.1 &92.7  & 94.4   \\
Double  & 83.3 & 83.3 &83.3        & 89.6 &80.2  &84.7                                & 96.9& 96.5 & 96.7  \\
Multiple& - & - &-                 & 95.8& 75.3 & 84.3                        & 97.3 &96.3  & 96.8  \\
\hline

All  &94.2 & 88.0 &91.0                 & 94.2 & 85.7    & 89.8                      &97.0  &95.7 & 96.3\\
\hline
\end{tabular}
\caption{\label{details-table} Precision, Recall and Micro-F1 scores of our model tested on different types of relations in multiple datasets. Note for some relation types the score is not available (denoted as ``-") because there is no such type in the dataset (see Table 2).}
\end{table*}

\section{Experiments}
This section describes the experimental process and best results while testing our methods on multiple public datasets. We performed overall comparison experiments with the baseline methods and completed more fine-grained analysis and comparison in different types of complex relations. Codes and more details can be found in Supplementary Materials.
\subsection{Experimental Settings}
We use Nvidia Tesla V100 32GB for training. The BERT model we use is [BERT-Base, Uncased]. Hyper-parameters are shown in Table~\ref{para-tab}. The optimizer is Adam~\cite{kingma2014adam}. Based on our problem formulation as described in Section~\ref{method}, our model actually fits a binary classifier to predict whether a triplet exists or not. Therefore, it actually gives a probability for each possible triplet. We still need a threshold to divide the positive and negative classes, and we set it as $0.5$ for balance. More details are shown in Supplementary Materials.
 Our codes are developed on OpenNRE~\cite{han2019opennre}. 
\subsection{Baseline and Evaluation Metrics}
As described above, Pre-trained Language Model (PLM) is so powerful that it may lead to unfairness in the comparison between our method and some old methods. Therefore, we chose some recent work (C-AGGCN~\cite{guo2019attention},
GraphRel2p~\cite{fu2019graphrel},
${\rm {BERT_{EM}}}$-MTB ~\cite{soares2019matching},
HBT~\cite{wei2019novel}) \textbf{published after the appearance of PLMs, especially BERT,} as our baseline. Such a selection is useful for measuring whether we have better exploited the potential of PLM in relational extraction tasks, rather than only relied on its power. As usual, we used the micro-F1 score as evaluation criteria.
\subsection{Datasets}
We performed our experiments on three commonly used public datasets (SemEval 2010 Task 8~\cite{hendrickx2009semeval}, NYT~\cite{riedel2010modeling},   WebNLG~\cite{gardent2017creating}) and compared the performance of our method to the baseline methods mentioned above. We followed splits and special process for the datasets conducted by the previous baseline models. More details are in Supplementary Materials.

We found that the complexity of the samples in these data sets varied widely. Similar to the  approach of Copy\_re ~\cite{zeng2018extracting}, we measured the complexity of the relations in the three datasets from two dimensions, i.e., \textbf{the number of samples with overlapping relations} and \textbf{the number of samples with multiple relations}. 

For overlapping relations, we followed the method proposed in Copy\_re~\cite{zeng2018extracting} to divide samples into three types: ``Normal" as all relations in the sample is normal; ``EPO" as there are at least two relations overlapped in the same entity-pair in the sample; ``SEO" as there are at least two relations sharing a single entity in the sample. These three types can reflect the complexity of relations. 
For multiple relations, we also divide samples into three types: ``Single" as only one relation appears in the sample, while ``Double" as two and ``Multiple" as no less than three. These three types can reflect the complexity of samples.

Table~\ref{data-table} shows the complexity analysis of each dataset.

\subsection{Results and Analysis}

Table~\ref{preformance-table} shows the performance of our method on all test data sets and the comparisons with the corresponding baseline methods. Given different test settings, we find that \textbf{our model generally outperformed the baseline models}, with a margin over the optimal baseline ranging from $1\%$ to $8\%$.

\begin{figure*}[ht]
\centering
\subfigure{
\begin{minipage}[t]{0.32\linewidth}
\centering
\includegraphics[width=1.8in]{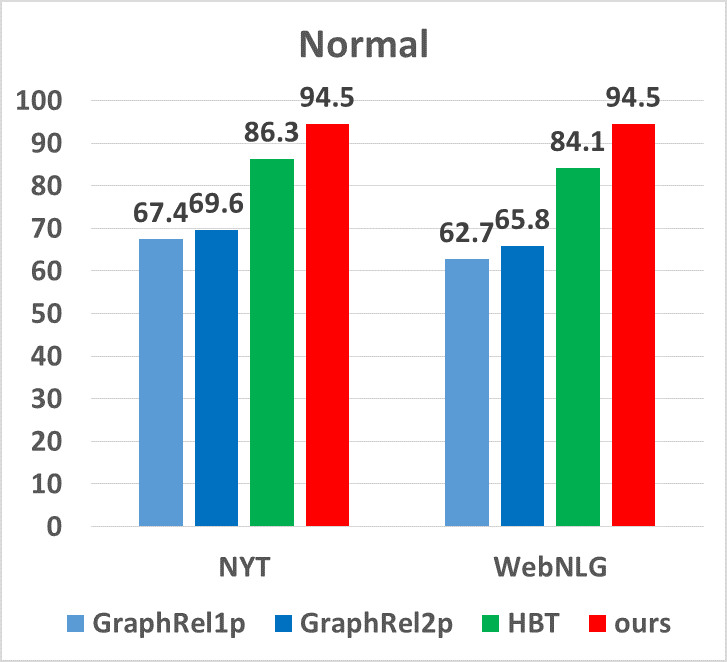}
\end{minipage}%
}%
\subfigure{
\begin{minipage}[t]{0.32\linewidth}
\centering
\includegraphics[width=1.8in]{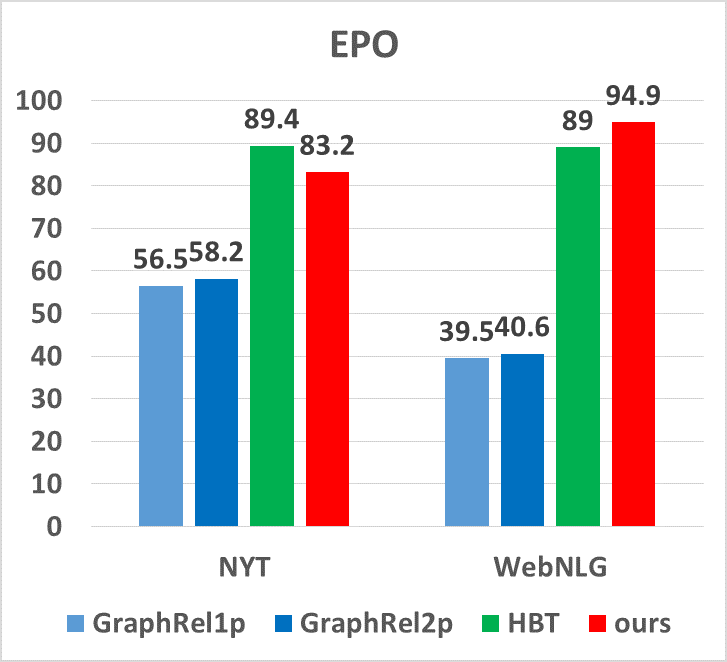}
\end{minipage}%
}%
\subfigure{
\begin{minipage}[t]{0.32\linewidth}
\centering
\includegraphics[width=1.8in]{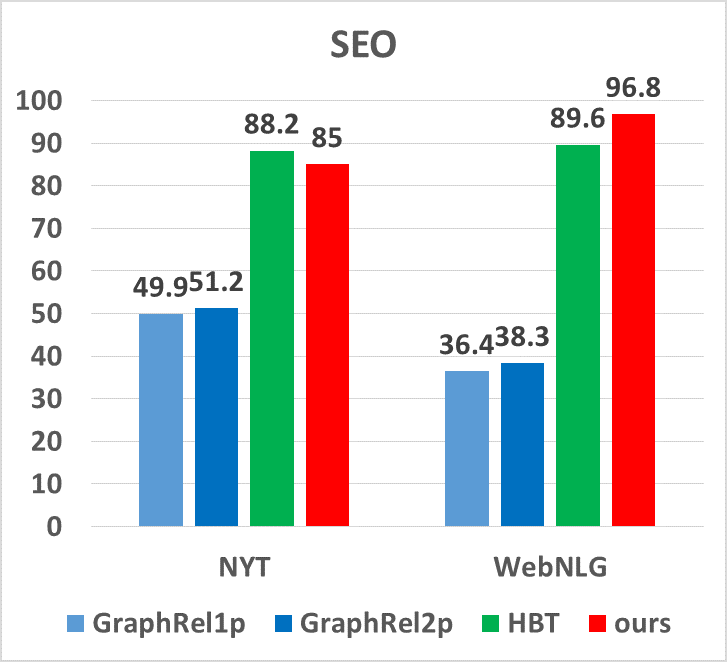}
\end{minipage}
}%
\centering
\caption{Micro-F1 scores of Normal, Entity Pair Overlapped (EPO) and Single Entity Overlapped (SEO). Our methods are tested on NYT and WebNLG, with comparison of GraphRel and HBT.}

\label{overlapped}
\end{figure*}

\begin{figure*}[htbp]
\centering
\subfigure{
\begin{minipage}[t]{1\linewidth}
\centering
\includegraphics[width=4in]{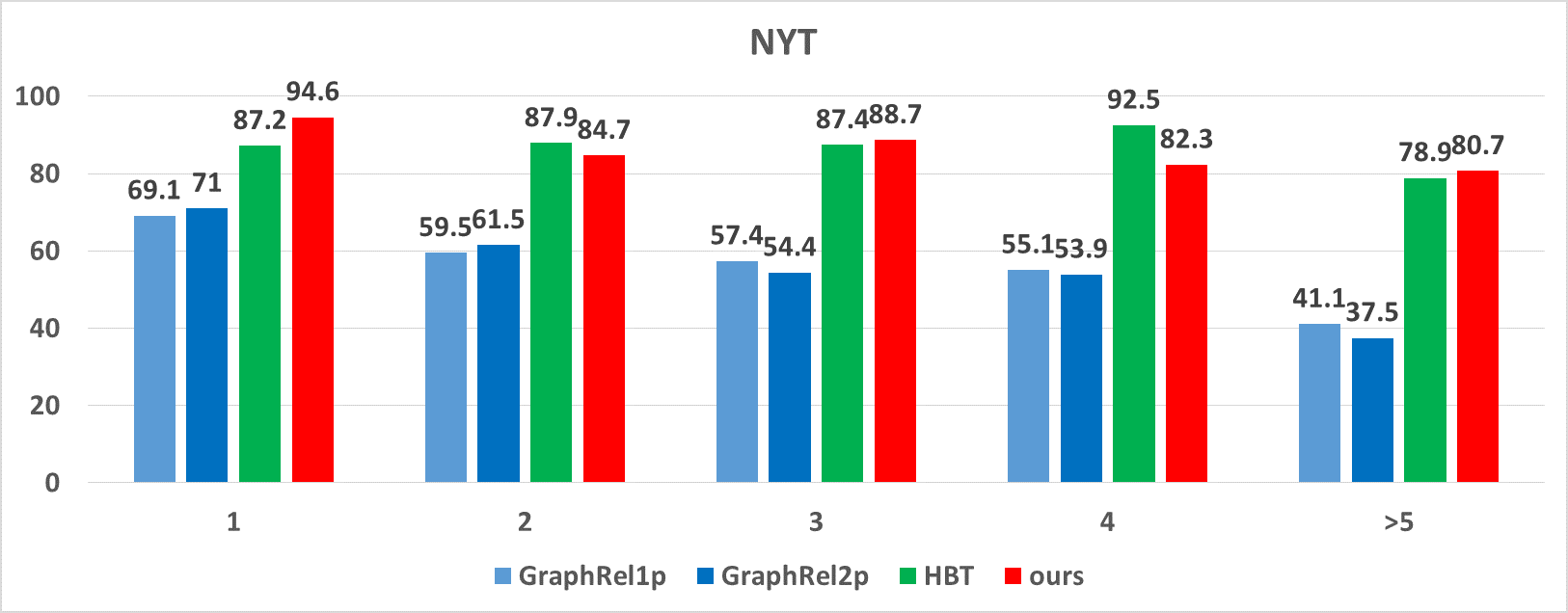}
\end{minipage}%
}%

\subfigure{
\begin{minipage}[t]{1\linewidth}
\centering
\includegraphics[width=4in]{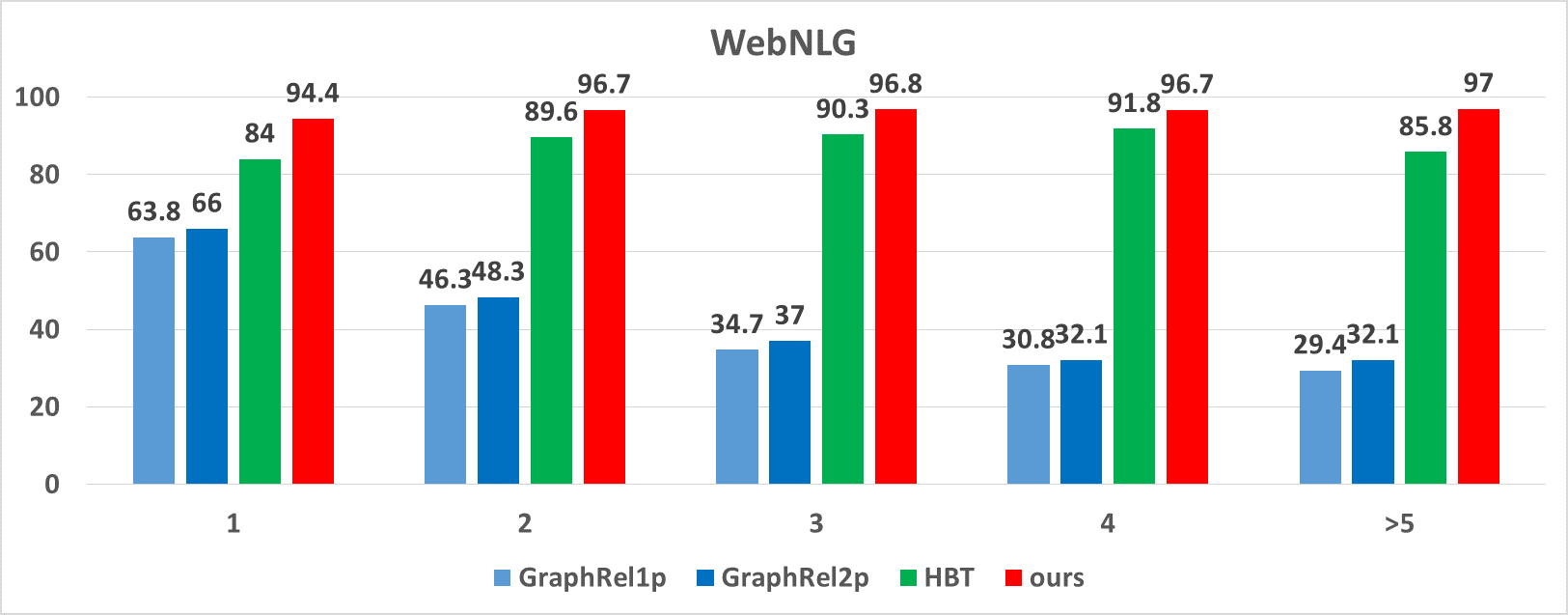}
\end{minipage}%
}%
\centering
\caption{Micro-F1 scores of $x$-relations in one sample sentence. (Here $x=1,2,3,4$ or $x\geq5$.) Our methods are tested on NYT and WebNLG, with comparison of GraphRel and HBT.}

\label{multi}
\end{figure*}

\begin{table*}[ht]
\small
\centering
\begin{tabular}{lcccc}
\hline
\multirow{2}*{}  & \multicolumn{2}{c}{\underline{\textbf{NYT}}}& \multicolumn{2}{c}{\underline{\textbf{WebNLG}}}\\
~&Complex& Simple &Complex& Simple\\
\hline
\multirow{10}{1cm}{Text}
& \multirow{10}{2.5cm}{\textbf{Ernst Haefliger}...died on Saturday in \textbf{Davos}, Switzerland, where he maintained a second home. } & \multirow{10}{2.5cm}{Georgia Powers...said \textbf{Louisville} was finally ready to welcome \textbf{Muhammad Ali} home.}  &
\multirow{10}{2.5cm}{The 1 Decembrie 1918 University is located in \textbf{Alba Iulia}, \textbf{Romania}. The capital of the country is \textbf{Bucharest}...} & \multirow{10}{2.5cm}{The Germans of \textbf{Romania} are one of the ethnic groups in Romania...the 1 Decembrie 1918 University is located in the city of \textbf{Alba Iulia}.} \\
&  & & \\
&  & & \\
&  & & \\
&  & & \\
&  & & \\
&  & & \\
&  & & \\
&  & & \\
&  & & \\
\hline
\multirow{3}{1cm}{Labels}
& \multirow{5}{2.5cm}{{\color{red}(place of birth:Ernst Haefliger, Davos)}, (place of death:Ernst Haefliger, Davos)  } & \multirow{5}{2.5cm}{(place of birth:Muhammad Ali, Louisville)}  &
\multirow{5}{2.5cm}{(is country of:Romania, Alba Iulia),
\\(is capital of:Bucharest, Romania)} & \multirow{5}{2.5cm}{(is country of:Romania, Alba Iulia)} \\
&  & & \\
&  & & \\
&  & & \\
&  & & \\
&  & & \\

\hline
\multirow{3}{1.5cm}{Predictions}
& \multirow{3}{2.5cm}{(place of death:Ernst Haefliger, Davos)  } & \multirow{3}{2.5cm}{(place of birth:Muhammad Ali, Louisville)}  &
\multirow{3}{2.5cm}{(is country of:Romania, Alba Iulia),
\\(is capital of:Bucharest, Romania)} & \multirow{3}{2.5cm}{{\color{red}(ethnic groups in:Romania, Alba Iulia)}} \\
&  & & \\
&  & & \\
&  & & \\
&  & & \\
&  & & \\
\hline
\end{tabular}

\caption{\label{samples} Examples from NYT and WebNLG to compare influences of semantic correlations while predicting simple and complex relations. Red triplets are fake relations.}
\end{table*}

To explain the difference in performance margin, we design a detailed experiment to evaluate how our model performs in each relation type as shown in Table~\ref{details-table}. On the other hand, we also conducted the comparison regarding each relation type with 3 baseline methods  ($\rm {GraphRel_{1p}}$~\cite{fu2019graphrel}, $\rm {GraphRel_{2p}}$ ~\cite{fu2019graphrel},
HBT~\cite{wei2019novel}) in Figure~\ref{overlapped} and Figure~\ref{multi}. We only compared F1 scores on NYT and WebNLG, since nearly no complex relations exist in SemEval. Detailed analyses on each dataset are listed as below.

\textbf{SemEval}. SemEval only has around $20$ samples with two relations (``Others" class excluded), with only $6$ of them in test dataset. Thus in Double-relation type, our model's performance crashes down about $8\%$ because of large variance. 

\textbf{NYT}. On NYT our method has experienced large fluctuations. The reason is that NYT is the only dataset constructed by distant supervision~\cite{mintz2009distant}, so the data quality is low. Distant supervision methods, which automatically label data from knowledge graphs, bring more errors in complex relations than simple relations. Therefore, although it seems that the amount of complex relations is sufficiently large, the performance of our model on complex relations still lags behind that on simple relations (around $10\%$ lower in F1 score, around $17\%$ lower in Recall). Nonetheless, comparison results still show that our metric scores while dealing with simple relations are higher than both baselines (around $7\%$ higher than HBT and $25\%$ higher than GraphRel). Even for complex relations, we are still significantly better (around $30\%$) than GraphRel, but a little bit lower (around $3\%$) than HBT.

\textbf{WebNLG}. It is easy to find our performances (Precision, Recall and Micro-F1 score) keep stable no matter how complicated the type is, except for EPO. It is because there are only around $100$ samples in EPO thus the model's performances are of high variance. Interestingly, the Micro-F1 scores of complex relations are even higher than simple relations around $2\%$, which is further discussed below. Comparison results show our model is far better than baselines (around $8\%$ higher than HBT and $50\%$ higher than GraphRel).

Given the results from all the datasets, our method shows consistent high performance on simple relation extraction tasks. Furthermore, it generally \textbf{demonstrates stable performance when faced with more challenging settings including overlapped relation and multiple relation extraction}.

Another interesting phenomenon is, on WebNLG, our model does better (around $1.5\%$ higher on F1 score, $3\%$ higher on Recall) while dealing with complicated relations than simple relations.  Our guess is that since our model can predict multiple relations at the same time, \textbf{it may combine semantic correlations between multiple relations to find more annotated relations by preventing some semantic drift}. On the other hand, on NYT, where semantic correlations between multiple relations generated by distant supervision are very likely to be fake, our model tends to neglect those meaningless semantic correlations. Therefore, it filters out potential falsely labeled relations and generate lower Recall.

To support the above reasoning, Table~\ref{samples} illustrates some real examples from WebNLG and NYT dataset. On WebNLG, our examples demonstrate the beneficial effects from properly labeled complex relations on our model. In the simple sample, our model made a mistake to consider “Romania” as an ethnic group in “Alba Iulia”, while in the complex sample from WebNLG, the model made the correct prediction that “Romania” is the country of “Alba Iulia”, by successfully identifying “Bucharest” as the capital of “Romania”. In comparison, for the simple sample from NYT, the model predicted ``place of birth" correctly, while failed to predict it in the complex sample, since this relation is not real and also has no semantic correlations with the correct relation ``place of death".

\section{Conclusion}

This paper introduces a downstream network architecture of pre-trained language model to process supervised relation extraction. The network calculates the relation score matrix of all entity pairs on all relation types by extracting the different head and tail entities' embeddings from the pre-trained language model. Experiments have shown that it has achieved significant improvements across multiple public datasets when compared to current best practices. Moreover, further experiments demonstrate the ability of this method to deal with complex relations. 
Also, we believe this network will not conflict with many other methods, thus it can be combined with them (e.g., use other special PLMs like ERNIE~\cite{zhang2019ernie}, ${\rm {BERT_{EM}}}$-MTB ~\cite{soares2019matching}) and performs better.

In addition, we believe that the current architecture has the potential to be improved for dealing with many other relation problems, including applications in long-tail relation extraction, open relation extraction, and joint extraction and so on. 

\bibliography{acl2020}
\bibliographystyle{plain}

\end{document}